\documentclass[runningheads]{llncs}
\usepackage{graphicx}
\usepackage{tikz}
\usepackage{comment}
\usepackage{amsmath,amssymb} % define this before the line numbering.
\usepackage{color}
\usepackage{orcidlink}

\usepackage{bm}
\usepackage{here}
\usepackage{multirow}

\newcommand{\bhline}[1]{\noalign{\hrule height #1}} % added for bold line in table
\DeclareMathOperator*{\argmin}{arg\,min} % defined for argmin in equations
\begin{document}
\pagestyle{headings}
\mainmatter
\def\ECCVSubNumber{1552}  % Insert your submission number here

\title{Revisiting a kNN-based Image Classification System with High-capacity Storage} % Replace with your title

\begin{comment}
\titlerunning{ECCV-22 submission ID \ECCVSubNumber} 
\authorrunning{ECCV-22 submission ID \ECCVSubNumber} 
\author{Anonymous ECCV submission}
\institute{Paper ID \ECCVSubNumber}
\end{comment}
%******************

%\begin{comment}
\titlerunning{Revisiting a kNN-based Image Classification System}
\author{Kengo Nakata \and
Youyang Ng \and
Daisuke Miyashita \and
Asuka Maki \and \\
Yu-Chieh Lin \and
Jun Deguchi
}
\authorrunning{K. Nakata et al.}
\institute{
 Kioxia Corporation, Kawasaki, Japan \\
\email{\{kengo1.nakata, youyang.ng, daisuke1.miyashita, asuka.maki, \\ yuchieh.lin, jun.deguchi\}@kioxia.com}}

% original ->
%\author{First E. van Author\inst{1}\orcidlink{0000-1111-2222-3333}\index{van Author, First E.} \and
%Second Author\inst{2,3}\orcidlink{1111-2222-3333-4444} \and
%Third Author\inst{3}\orcidlink{2222--3333-4444-5555}}
%% First names are abbreviated in the running head.
%% If there are more than two authors, 'et al.' is used.
%%
%\institute{Princeton University, Princeton NJ 08544, USA \and
%Springer Heidelberg, Tiergartenstr. 17, 69121 Heidelberg, Germany
% <- original
%\end{comment}
%******************

\maketitle

\begin{abstract}
In existing image classification systems that use deep neural networks, the knowledge needed for image classification is implicitly stored in model parameters.
If users want to update this knowledge, then they need to fine-tune the model parameters.
Moreover, users cannot verify the validity of inference results or evaluate the contribution of knowledge to the results. 
In this paper, we investigate a system that stores knowledge for image classification,
such as image feature maps, labels, and original images, not in model parameters but in external high-capacity storage.
Our system refers to the storage like a database when classifying input images.
To increase knowledge, our system updates the database instead of fine-tuning model parameters,
which avoids catastrophic forgetting in incremental learning scenarios.
We revisit a kNN (k-Nearest Neighbor) classifier and employ it in our system.
By analyzing the neighborhood samples referred by the kNN algorithm, we can interpret how knowledge learned in the past is used for inference results.
Our system achieves 79.8\% top-1 accuracy on the ImageNet dataset without fine-tuning model parameters after pretraining,
and 90.8\% accuracy on the Split CIFAR-100 dataset in the task incremental learning setting.
\end{abstract}

\section{Introduction}
\label{sec:introduction}

Image classification systems using deep neural networks (DNN) have achieved superhuman recognition performance in computer vision tasks~\cite{LeNet,AlexNet,VGG,ResNet,ViT,CoAtNet}.
On the other hand, the knowledge for image classification is implicitly stored in model parameters and is not accessible to users.
For example, users cannot retrieve knowledge related to only \textit{cats} or \textit{dogs} from the model parameters.
Training datasets may contain a small amount of inappropriate data
(e.g., incorrectly labeled or undesirably biased images)~\cite{ImageNetLabelError,ImageNetReaL,UncoveringBiases}.
If the model parameters are optimized using such training datasets, then the image classification systems will implicitly contain false knowledge.
In existing systems, however, users cannot selectively eliminate or correct only the false knowledge in model parameters.

To update knowledge (add, delete, or modify), users need to fine-tune the model parameters.
Especially, in continual or incremental learning scenarios~\cite{SurveyIncLearn,SurveyOnlineLearn} where new data is continually added to a training dataset,
the cost for fine-tuning is incurred repeatedly as new data is added.
If model parameters are fine-tuned with only newly added data in an attempt to reduce the cost,
the model will acquire new knowledge for the added data while forgetting the knowledge learned in the past,
which is often called catastrophic forgetting~\cite{cataforget,SurveyIncLearn}.
Although various methods have been proposed to mitigate the impact of catastrophic forgetting~\cite{iCaRL,GEM,AGEM,EWC,ER,ERTrick},
as long as modifying model parameters, forgetting some knowledge is inevitable.
Refs.~\cite{CLIP,ALIGN} propose a zero-shot classifier that does not require any fine-tuning 
by sufficiently pretraining model parameters on large-scale datasets which contain a variety of images.
Such classifier does not face the catastrophic forgetting even without fine-tuning, but it cannot additionally acquire user's desired knowledge with the user's own datasets.

Moreover, in existing systems, users cannot verify the validity of inference results or evaluate the contribution of knowledge to the results.
For instance, users do not know how models utilize knowledge learned in the past to classify input images.
Although we can identify unnecessary parameters through accuracy evaluation or some other parameter analyses~\cite{bayesian_pruning,reinforcement_pruning,AdvPrune}, 
we cannot interpret why the parameters, that is, knowledge, are unnecessary (cf. explainable AI).

We investigate a system that stores image feature maps, labels, and original images of entire training datasets in external high-capacity storage as knowledge for image classification.
Our image classification system refers to the external storage like a database when classifying input images.
To increase knowledge, our system adds image feature maps and labels for new data to the database instead of fine-tuning model parameters using those new data. 
We employ a kNN (k-Nearest Neighbor) classifier~\cite{kNNreview}, which is one of the most classic classification algorithms.
Recently, many papers have employed kNN for the purpose of evaluating their proposed representation learning algorithms~\cite{DINO,iBOT,EsViT}.

In this paper, we shed new light on the potentials of kNN classification system with high-capacity storage.
A concurrent work~\cite{longtail_retrieval} demonstrates that kNN retrieval improves long-tail recognition.
Not only this, we empirically show that million scale kNN retrieval is affordable for practical image classification tasks,
and that our system with kNN avoids catastrophic forgetting in continual learning scenarios and achieves better accuracy than conventional methods.
Furthermore, by reviewing neighborhood samples referred by the kNN algorithm, we can verify the validity of inference results,
such as whether those referred samples contain incorrectly labeled data.
If those referred samples contain incorrectly labeled data,
our system can correct the false knowledge and improve the accuracy by eliminating only the incorrectly labeled data from the database.

The main contributions of this paper are as follows:
\begin{itemize}
\item We investigate a large-scale kNN system that stores knowledge for image classification in high-capacity storage and refers to the storage when classifying images, and empirically demonstrate its effectiveness and applicability on various image classification datasets and in continual learning scenarios.
\item We also show that a large-scale kNN system has a capability of verifying the validity of inference results and selectively correcting only false knowledge.
\end{itemize}

\section{Related Work}
\label{sec:related_works}

\subsection{Data-driven Image Classification}
\label{sec:rw_knowledge}

DNNs, hundreds of millions to billions parameters~\cite{ViT,CCT,CoAtNet,PyramidViT}
of which are optimized by supervised learning with labeled images, have achieved state-of-the-art results
in visual understanding and image classification tasks.
In recent years, unsupervised visual clustering, which does not rely on a labeled dataset for learning 
the feature representation and classifier, has been developed~\cite{ID,SimCLR,MOCO,MOCOV2,CLD,MAE,NNCLR}.
Unsupervised pretraining strategies on large-scale datasets that include unlabeled or noisily labeled images have demonstrated
the potential of large-scale open-domain knowledge for improving the accuracy of closed-domain image classification tasks~\cite{ViT,ALIGN,CLIP}.
In addition, although the applicable dataset scale is still limited, Ref.~\cite{datapoints_attention} presented an architecture that utilizes an entire dataset (e.g., all training images) to classify an input image during the inference process.
Data has never been more important in the quest to further improve the usability of image classification.
Inspired by the data-driven approaches in classification strategies, 
our work focuses on the application of knowledge retrieval to image classification.
We present a method
that utilizes the potential of both the trained parametric representation model and the available 
datasets during the inference process.

\subsection{Knowledge Retrieval}
\label{sec:rw_knowledge}

Knowledge retrieval has seen substantial advancement in recent years, particularly in DNN-based natural language processing (NLP).
DPR~\cite{DPR} applied a dense representation to passage retrieval in open-domain question answering tasks
and achieved better retrieval performance than traditional sparse vector space
models, such as TF-IDF and BM25. KEAR~\cite{hummancommonsense} brought external knowledge into the predicting 
process of Transformer~\cite{Transformer} to reach human parity in a challenging commonsense task~\cite{CommonsenseQA}.
RETRO~\cite{RETRO} introduced a frozen kNN retriever into the
Transformer architecture in the form of chunked cross-attention to enhance the performance of 
auto-regressive language models. External world knowledge has been retrieved to assist in solving
various NLP tasks. Our work looks to extend the adoption of knowledge retrieval beyond the modality of NLP.
We introduce an image classification architecture based on knowledge retrieval, which is a data-driven paradigm that is centralized
on available large-scale data resources and is supported by a trained representation model and a kNN classifier.

\subsection{Continual Learning}
\label{sec:rw_increment}

When new data or tasks are added continually after a training process,
fine-tuning is used to update the previously trained models with new knowledge. 
This continual learning, also known as incremental learning, online learning, or lifelong learning~\cite{SurveyIncLearn,SurveyOnlineLearn}, is similar to natural human learning processes 
and is a key challenge for achieving artificial general intelligence.
However, the simple procedure of iterative fine-tuning using only new knowledge suffers from catastrophic forgetting~\cite{cataforget,SurveyIncLearn}.
To mitigate the impact of catastrophic forgetting, 
GEM~\cite{GEM} introduced episodic memory to store a subset of the previously learned data,
enabling an external memory-driven continual learning strategy.
ER~\cite{ER}, inspired by the human brain's ability to replay past experiences, introduced the experience replay of memory.
Episodic memory and experience replay have seen great developments in recent years~\cite{AGEM,ERTrick,HAL}. 
iCaRL~\cite{iCaRL} evaluated the effect of catastrophic forgetting under a class incremental learning setting,
where new classes are incrementally added to classifiers, 
and proposed a rehearsal approach with exemplar images of old classes being stored in memory. 
Inspired by the aforementioned works in adopting memory and the replay mechanism, we incorporate direct knowledge 
retrieval to solve the catastrophic forgetting problem in continual learning scenarios. Instead of a model-based approach, we introduce 
a data-based approach, leveraging available datasets as knowledge sources to adapt to the incrementation of tasks and classes.

\subsection{Explainable AI}
\label{sec:rw_explainable}

DNN models have been a technological breakthrough for various computer vision tasks.
However, the explainability of DNN models remains a challenge and has led to
slower-than-expected deployments in critical infrastructures.
Attempts~\cite{CAM,GradCAM,TIBAV} have been made to interpret image classification models.
GradCAM~\cite{GradCAM} proposed a gradient-based localization technique for rendering attention
maps on the input images. Ref.~\cite{TIBAV} went beyond attention visualization by adopting gradients and the propagation of relevancy scores.
These analysis methods can visualize the areas of the input images that DNN models focus on during the inference process,
but they cannot analyze how the DNN models use knowledge acquired in training when classifying the input image.
For example, if a training dataset contains false data (e.g., incorrectly labeled images),
then these methods will not be able to selectively retrieve knowledge related to the false data from the trained models nor evaluate the impact of the false knowledge on the classification results.
Our data-driven image classification architecture visualizes how knowledge is used to classify input images and enables selective modifications to specific knowledge without fine-tuning.

\begin{figure}[t]
   \begin{center}
   \includegraphics[width=0.85\linewidth]{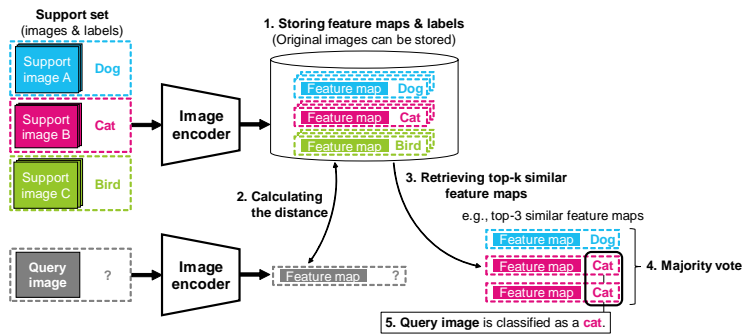}
   \end{center}
   \caption{Overview of our image classification system. Our system stores feature maps extracted from support images with the corresponding labels to the external storage.
   When classifying a query image, our system retrieves feature maps similar to the query one from the storage by calculating the distance based on cosine similarity.
   The query image is classified by majority vote on the labels of the top-k similar feature maps.}
   \label{fig:overview}
\end{figure}

\section{Approach}
\label{sec:proposed_system}

Fig.~\ref{fig:overview} provides an overview of our image classification system.
Our system has three phases: pretraining, knowledge storing, and inference.
The first step is to pretrain an image encoder model on a large-scale dataset containing a variety of images.
This pretraining can use unlabeled or noisily labeled datasets.
In the knowledge storing phase, the pretrained image encoder model extracts feature maps
from support sets (e.g., a training dataset of a user's desired downstream task).
Extracted feature maps are paired with their corresponding labels and registered in the external storage like a database.
In the inference phase, the pretrained image encoder model extracts a feature map of a query image.
Our system retrieves feature maps similar to the query one from the storage by calculating the distance based on cosine similarity.
Then, the query image is classified by majority vote on the paired labels of the top-k similar feature maps retrieved.
The following sections describe each phase in detail.

\subsection{Pretraining}
\label{sec:pm_pretraining}

We pretrain an image encoder model to learn feature representations.
We assume that this process is conducted on a large-scale computer system, such as a data center (not on the user side)~\cite{foundation_model}.
Similarity-based retrieval in the latent space is performed in the following inference phase (see Section~\ref{sec:pm_inference} for details).
Therefore, transformations by the pretrained image encoder models should be such that semantically similar images are mapped to the neighborhood in the latent space.
In this section, we describe suitability of pretraining methods that can be employed in our system.

\noindent
\textbf{Supervised learning with specific datasets.}
Models trained on a specific dataset by supervised learning can be used for our system. % Models pretrained
An example is a ResNet-50 model~\cite{ResNet} trained on the ImageNet-1k classification task~\cite{ImageNet}.
If input images are classified as the first class (e.g., tench) by the trained model, those input images are mapped to the feature maps so that
they have a high similarity to the vector of weight parameters corresponding to the first class in the last fully connected layer.
Therefore, we can employ the trained model as an image encoder by excluding the last fully connected layer.
However, the trained model can overfit the classification setting of the specific dataset (e.g., the 1000 classes of ImageNet-1k).

\noindent
\textbf{Self-supervised contrastive learning.}
We can also employ self-supervised contrastive learning methods (e.g., SimCLR~\cite{SimCLR} and MoCo~\cite{MOCO,MOCOV2}), % CLD~\cite{CLD}
which learn image feature representations based on the similarity or dissimilarity of images between two or more views.
Among the contrastive learning methods, CLIP~\cite{CLIP} and ALIGN~\cite{ALIGN} 
jointly learn image and text feature representations of image and text encoders using a large number of image and text pairs collected from the internet.
These methods do not require hand-crafted labeling or annotation, and the number of semantic labels is much larger than the number of labels on commonly used image datasets, such as the limited 1000 labels in ImageNet-1k. % annotation costs

\noindent
\textbf{Other self-supervised learning.}
Masked Auto Encoder~\cite{MAE} learns image feature representations through a task of reconstructing the original images from masked images. 
By fine-tuning on the labeled dataset, the encoders pretrained with Masked Auto Encoder have achieved high accuracy in the ImageNet-1k classification task.
Because this pretraining method is label-independent, the pretrained model seems unlikely to overfit a specific classification setting (e.g., the 1000 classes of ImageNet-1k).
However, the objective function of this pretraining method does not explicitly make the encoder model to map semantically similar images to the neighborhood in the latent space,
so it is not suitable for similarity-based retrieval in our system.

\noindent
\textbf{Selecting pretraining methods.}
We conducted a preliminary experiment using three pretrained image encoder models:
(1) a ResNet-50 model trained on the ImageNet-1k dataset with the labels by supervised learning,
(2) a Vision Transformer~\cite{ViT} Base model with input patch sizes of 16$\times$16 (ViT-B/16) trained by CLIP on 400 million image and text pairs collected from the internet,
(3) a ViT-B/16 model trained by Masked Auto Encoder (MAE) on the ImageNet-1k dataset without the labels.
We employed the three pretrained image encoders in our system, and evaluated the test accuracy on various image datasets (CIFAR-10, CIFAR-100~\cite{CIFAR10}, STL-10~\cite{STL10}, and ImageNet-1k).

We summarize the test accuracy in Table~\ref{table:pretrained_comparison}.
As discussed above, (1) the supervised learned model on ImageNet-1k achieves the best accuracy on ImageNet-1k, but not on the other datasets.
This supervised learned model does not generalize well to unseen datasets. 
(3) the model pretrained by MAE shows poor performance since the objective of this pretraining method is not compatible with similarity-based retrieval in our system.
On the other hand, (2) the model pretrained by CLIP achieves good accuracy on various datasets, indicating that the model is well generalized.
Based on the result of this preliminary experiment, we employed image encoder models pretrained by CLIP in the experiments of Section~\ref{sec:experiments}.
The exploration of better pretraining methods is our future work.

\begin{table}[t]
\begin{center}
\caption{Accuracy comparison using different pretrained image encoder models.}
\scalebox{0.85}{
\begin{tabular}{ccc c c c}
& & \textbf{ImageNet-1k} & \textbf{CIFAR-10} & \textbf{CIFAR-100} & \textbf{STL-10} \\ \bhline{1.25pt} 
\multirow{2}{*}{(1)} & Supervised learned & \multirow{2}{*}{\textbf{74.9}} & \multirow{2}{*}{85.9} & \multirow{2}{*}{64.7} & \multirow{2}{*}{96.7} \\
& on ImageNet-1k & & & & \\ \hline
(2) & CLIP~\cite{CLIP} & 74.0 & \textbf{94.4} & \textbf{74.3} & \textbf{98.9} \\ \hline
(3) & MAE~\cite{MAE} & 26.7 & 51.8 & 23.2 & 66.6 \\ \hline
\end{tabular}
}
\label{table:pretrained_comparison}
\end{center}
\end{table}

\subsection{Knowledge Storing}
\label{sec:pm_storing}

To acquire knowledge, a pretrained image encoder model extracts feature maps from a support set, such as a training dataset of a user's desired downstream task.
Given a support set of $n$-labeled images $\{\bm{x}_{s,1}, ... , \bm{x}_{s,n}\}$, the corresponding labels $\{y_1, ... , y_n\}$, and the pretrained image encoder model $f(\cdot)$,
the $i$-th support image is mapped to the $d$-dimensional latent space, and the extracted feature map $\bm{z}_{s,i}$ is obtained by $\bm{z}_{s,i} = f(\bm{x}_{s,i}), \bm{z} \in \mathbb{R}^d$.

Unlike fine-tuning, our system does not require iterative forward, backward and parameter-update operations, but it just requires one forward operation of each image in the support set for the feature extraction.
Thus, the cost and effort are less than those of fine-tuning.
Moreover, our system can avoid catastrophic forgetting without fine-tuning, even when the knowledge is iteratively updated in continual learning scenarios.

\begin{figure}[t]
   \begin{center}
   \includegraphics[width=0.9\linewidth]{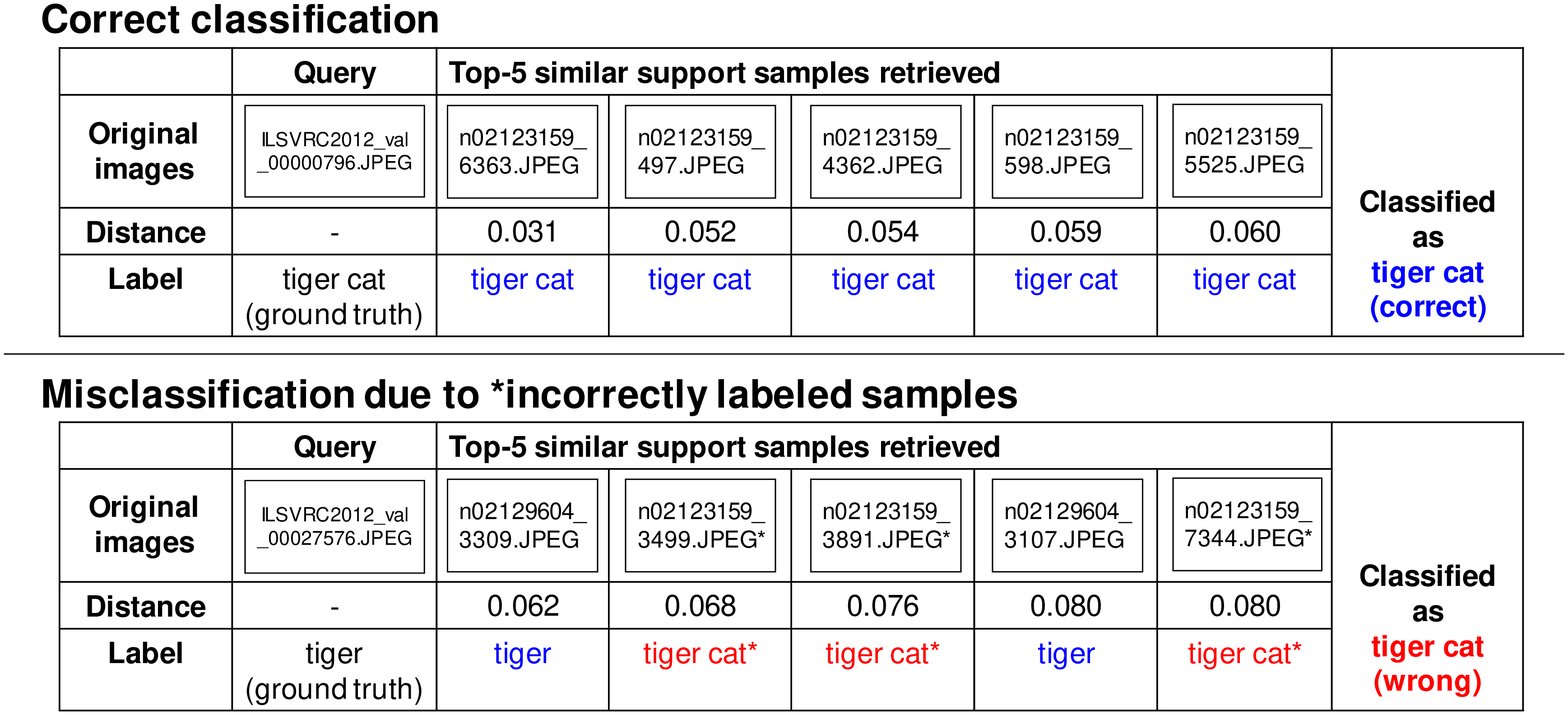}
   \end{center}
   \caption{Examples of query and the top-5 similar support samples retrieved.
   In these examples, validation and training data of ImageNet-1k are used for the query and the support set, respectively (the file names are also given in the examples).
   The top figure shows an example of \textcolor{blue}{correct classification}, and the bottom figure shows one of \textcolor{red}{misclassification} due to *incorrectly labeled support samples.
   By reviewing the original images and labels, we can verify the validity of the classification results.}
   \label{fig:retrieved_features}
\end{figure}

An extracted feature map is paired with the corresponding label such as $(\bm{z}_{s,i}, y_i)$, and registered in the external storage like a database.
If the original images are also registered to the database, then they can be used to verify the validity of classification results
(see Fig.~\ref{fig:retrieved_features} and Section~\ref{sec:pm_inference} for details).

\subsection{Inference}
\label{sec:pm_inference}

Our system classifies query images (i.e., inference) by referring to image feature maps and labels registered in the database.
The pretrained encoder first extracts a feature map $\bm{z}_q$ of a query image $\bm{x}_q$ as $\bm{z}_q = f(\bm{x}_q)$.
Our system retrieves support feature maps that are similar to query one from the database
by calculating the distance ($D$) based on cosine similarity between query and support feature maps as follows,
\begin{equation}
\label{eq:cos_sim}
D(\bm{z}_q,\bm{z}_{s,i}) = 1 - \frac{\bm{z}_q \cdot \bm{z}_{s,i}}{||\bm{z}_q||||\bm{z}_{s,i}||}.
\end{equation}
With $\argmin_{i \in n}D(\bm{z}_q,\bm{z}_{s,i})$, we can retrieve the index of the nearest, that is, the most similar support feature map to the query one.
Our system retrieves the index of top-k minimum distance, and the query image is classified based on the majority vote of the paired labels in order to mitigate the effect of outliers.

Fig.~\ref{fig:retrieved_features} shows examples of query and the retrieved samples from the support set.
The original images in the support set are not required for classifying query images.
However, the original images registered in the database can be used for evidence or for verification of the inference results.
For instance, as shown in Fig.~\ref{fig:retrieved_features}, not only the labels and distance values of retrieved feature maps,
but also the original images, can be listed as inference logs.
By reviewing the logs, we can verify that the retrieved samples contain inappropriate data (e.g., incorrectly labeled images).
The bottom figure in Fig.~\ref{fig:retrieved_features} shows an example of the misclassification
of a query image due to incorrectly labeled images in the support set (using a training dataset of ImageNet-1k).
If incorrectly labeled images are obtained in the logs, then we can correct false knowledge in our system 
by fixing the incorrect labels or eliminating such samples from the database.

The above characteristics are not present in existing image classification systems.
To correct false knowledge, existing systems need to fine-tune classifier models again after fixing incorrect labels in the support set.
On the other hand, our system does not require fine-tuning, thus eliminating the cost and effort for fine-tuning,
and can contribute to improving the explainability of AI by visualizing and reviewing referred images and labels when classifying input images.

\section{Experiments}
\label{sec:experiments}

We implemented our image classification system using the PyTorch library~\cite{pytorch}.
The setup and results for each experiment are described in detail below.

\begin{figure}[t]
   \includegraphics[width=1.0\linewidth]{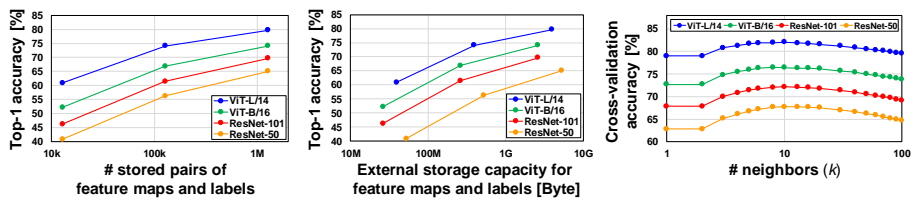}
   \caption{\textbf{Left/center}: the relation between the top-1 accuracy on the ImageNet-1k dataset and the number of stored pairs of feature maps and labels/the external storage capacity for storing the pairs.
            \textbf{Right}: the relation between cross-validation accuracy and the number of neighbors ($k$) for kNN.
            For cross-validation, the 1.28M training images in ImageNet-1k are randomly split 9:1 into support and query sets.}
   \label{fig:accuracy_stored_feature_topk}
\end{figure}

\subsection{Basic Performance Evaluation}
\label{exprmnt:basic_evaluation}

\noindent
\textbf{Experimental setup.} We evaluated the basic performance of our system using the CIFAR-10, CIFAR-100, STL-10, and ImageNet-1k datasets.
We employed image encoder models pretrained by CLIP.
We used ResNet-50 and 101 as CNN models and ViT-B/32, B/16, and L/14 as Vision Transformer models\footnote{ViT-B/32, B/16, and L/14 correspond to
Vision Transformer Base, Base, and Large model, with input patch sizes of 32$\times$32, 16$\times$16, and 14$\times$14, respectively.} for the image encoders.
These pretrained models extracted feature maps from training images, and those extracted feature maps and labels were registered in the database as support sets.
With the support sets, we evaluated the test accuracy of our system using test datasets as query sets.
As a baseline, we employed the test accuracy of the zero-shot CLIP classifier~\cite{CLIP}. This classifier can be applied to those classification tasks without any fine-tuning like our system.

In addition to the accuracy evaluation, we measured the processing time for inference of the ImageNet-1k images.
We used a NVIDIA A100 GPU and measured the processing time of our system and the zero-shot CLIP classifier as a baseline.
In this measurement, both the parameters of the image encoder model and the feature maps of the support set were loaded into the GPU memory from the external storage prior to the inference.
We then measured the processing time for transferring query images to the GPU, encoding the query images, and calculating the distance between the query and support feature maps.
The batch size was set to 1000, and the average processing time per image was calculated.\\

\noindent
\textbf{Experimental results.} The left graph in Fig.~\ref{fig:accuracy_stored_feature_topk} shows the relation between the top-1 accuracy on the ImageNet-1k dataset
and the number of stored pairs of feature maps and labels.
Here, we randomly sample images from the training dataset, adjusting the number of the stored pairs to 12.8k, 128k, and 1.28M.
We set the number of neighbors ($k$) for kNN to 10.
The more stored pairs that there are in the database, the better the accuracy will be that our system can achieve.
Moreover, the larger the model size, the better the performance of the feature extractor and also the accuracy.

The center graph in Fig.~\ref{fig:accuracy_stored_feature_topk} shows the relation between the top-1 accuracy and the external storage capacity for storing feature maps and labels.
In this experiment, the maximum storage capacity is 5.3 GB to store the feature maps and labels for 1.28M samples of ImageNet-1k.
As shown in the center graph, the larger the capacity, the better the accuracy.
% As shown in the center graph, the larger the data size using high-capacity storage, the better the accuracy.

In the right graph of Fig.~\ref{fig:accuracy_stored_feature_topk}, we evaluated the relation between the accuracy and the number of neighbors of $k$ by cross-validation.
For cross-validation, the 1.28M training images are randomly split 9:1 into support and query sets. % In this evaluation,
As shown in the right graph, the best accuracy is achieved when $k$ is set to around 10.
In the following experiments, $k$ is set to 10 unless otherwise noted.

\begin{figure}[t]
   \includegraphics[width=0.99\linewidth]{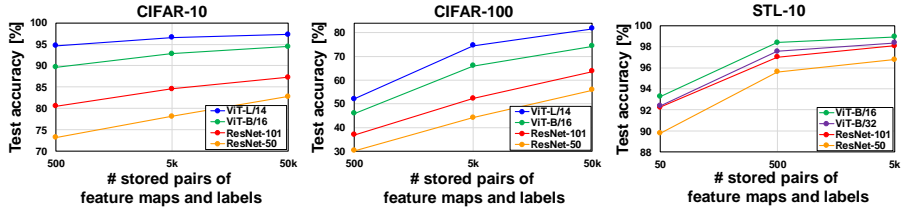}
   \caption{Relation between the test accuracy and the number of stored pairs of feature maps and labels for various image datasets.}
   \label{fig:accuracy_stored_feature}
\end{figure}

Fig.~\ref{fig:accuracy_stored_feature} shows the relation between the test accuracy and the number of stored pairs of feature maps and labels on the other datasets.
For all datasets, our system improves the test accuracy as the number of stored pairs increases
and as the model size becomes larger.

Table~\ref{table:summary_accuracy} summarizes the test accuracy of our system.
Our system can improve accuracy from the baseline (pretrained but not fine-tuned models) on the various image datasets
by storing image feature maps and labels of training datasets.

\begin{table}[t]
\caption{Test accuracy evaluations for various image datasets. Baseline indicates the accuracy of the zero-shot CLIP classifier.}
\scalebox{0.63}{
\begin{tabular}{ccc|ccc||cc|ccc||cc|ccc||cc|ccc|}
\hline 
\multicolumn{1}{|c}{}&\multicolumn{5}{|c||}{\textbf{CIFAR-10}}               &\multicolumn{5}{c||}{\textbf{CIFAR-100}} & \multicolumn{5}{|c||}{\textbf{STL-10}} & \multicolumn{5}{c|}{\textbf{ImageNet-1k}} \\ \cline{2-21}
\multicolumn{1}{|c}{}&\multicolumn{2}{|c}{ResNet}&\multicolumn{3}{|c||}{ViT} &\multicolumn{2}{c}{ResNet}&\multicolumn{3}{|c||}{ViT} &\multicolumn{2}{c}{ResNet}&\multicolumn{3}{|c||}{ViT} &\multicolumn{2}{c}{ResNet}&\multicolumn{3}{|c|}{ViT} \\ 
\multicolumn{1}{|c}{}&\multicolumn{1}{|c}{50}   &  101  & B/32 & B/16 & L/14                   &  50  &  101  & B/32 & B/16 & L/14 & 50 &  101  & B/32 & B/16 & L/14                   &  50  &  101  & B/32 & B/16 & L/14\\ \hline 
\multicolumn{1}{|c}{\textbf{Baseline}} &\multicolumn{1}{|c}{75.6} & 81.0  & 91.3 & 91.6 & 96.2                   & 41.6 & 49.0  & 65.1 & 68.7 & 77.9 & 94.3 & 96.7  & 97.2 & 98.2 & 99.3                   & 59.6 & 62.2  & 63.2 & 68.6 & 75.3\\
\multicolumn{1}{|c}{\textbf{Ours}}     &\multicolumn{1}{|c}{82.8} & 87.3  & 92.7 & 94.4 & 97.3                   & 55.7 & 63.6  & 71.5 & 74.3 & 81.7  & 96.8 & 98.1  & 98.4 & 98.9 & 99.6                   & 65.0 & 69.7  & 67.0 & 74.0 & 79.7\\
\multicolumn{1}{|c}{$\Delta$} &\multicolumn{1}{|c}{\textbf{+7.2}} & \textbf{+6.3} & \textbf{+1.4} & \textbf{+2.8} & \textbf{+1.1} & \textbf{+14.1} & \textbf{+14.6} & \textbf{+6.4} & \textbf{+5.6} & \textbf{+3.8} & \textbf{+2.5} & \textbf{+1.4} & \textbf{+1.2} & \textbf{+0.7} & \textbf{+0.3} & \textbf{+5.4} & \textbf{+7.5} & \textbf{+3.8} & \textbf{+5.4} & \textbf{+4.4}\\ \hline
\end{tabular}
}
\label{table:summary_accuracy}
\end{table}

\begin{table}[t]
\caption{Accuracy comparison with and without data augmentation (DA) on the support sets.}
\scalebox{0.65}{
\begin{tabular}{ccc|ccc||cc|ccc||cc|ccc||cc|ccc|}
\hline 
\multicolumn{1}{|c}{}&\multicolumn{5}{|c||}{\textbf{CIFAR-10}}               &\multicolumn{5}{c||}{\textbf{CIFAR-100}} & \multicolumn{5}{|c||}{\textbf{STL-10}} & \multicolumn{5}{c|}{\textbf{ImageNet-1k}} \\ \cline{2-21}
\multicolumn{1}{|c}{}&\multicolumn{2}{|c}{ResNet}&\multicolumn{3}{|c||}{ViT} &\multicolumn{2}{c}{ResNet}&\multicolumn{3}{|c||}{ViT} &\multicolumn{2}{c}{ResNet}&\multicolumn{3}{|c||}{ViT} &\multicolumn{2}{c}{ResNet}&\multicolumn{3}{|c|}{ViT} \\ 
\multicolumn{1}{|c}{}&\multicolumn{1}{|c}{50}   &  101  & B/32 & B/16 & L/14                   &  50  &  101  & B/32 & B/16 & L/14 & 50 &  101  & B/32 & B/16 & L/14                   &  50  &  101  & B/32 & B/16 & L/14\\ \hline 
\multicolumn{1}{|c}{\textbf{w/o DA}}     &\multicolumn{1}{|c}{82.8} & 87.3  & 92.7 & 94.4 & 97.3                   & 55.7 & 63.6  & 71.5 & 74.3 & 81.7 &\multicolumn{1}{|c}{96.8} & 98.1  & 98.4 & 98.9 & 99.6                   & 65.0 & 69.7  & 69.0 & 74.0 & 79.7 \\
\multicolumn{1}{|c}{\textbf{w/ DA}}        &\multicolumn{1}{|c}{83.6} & 87.8  & 93.0 & 94.3 & 97.3                   & 56.8 & 63.9  & 71.8 & 74.7 & 81.7 &\multicolumn{1}{|c}{97.0} & 97.9  & 98.3 & 99.1 & 99.6                   & 65.4 & 70.2  & 69.2 & 74.3 & 79.8 \\
\multicolumn{1}{|c}{$\Delta$} &\multicolumn{1}{|c}{\textbf{+0.8}} & \textbf{+0.5} & \textbf{+0.3} & -0.1 & 0.0 & \textbf{+1.1} & \textbf{+0.3} & \textbf{+0.3} & \textbf{+0.4} & 0.0 &\multicolumn{1}{|c}{\textbf{+0.2}} & -0.2 & -0.1 & \textbf{+0.2} & 0.0 & \textbf{+0.4} & \textbf{+0.5} & \textbf{+0.2} & \textbf{+0.3} & \textbf{+0.1} \\ \hline
\end{tabular}
}
\label{table:data_augmentation}
\end{table}

Based on the experimental results in Figs.~\ref{fig:accuracy_stored_feature_topk} and~\ref{fig:accuracy_stored_feature},
we utilize a data augmentation technique to increase the number of stored pairs of feature maps and labels.
Data augmentation is typically used to prevent models from overfitting during training and to improve the generalization performance.
In particular, we use only horizontal flipping to augment support images (2$\times$), and set the number of neighbors of $k$ to 20.
Table~\ref{table:data_augmentation} summarizes the accuracy with and without data augmentation.
As shown in the results for CIFAR-100 and ImageNet-1k in Table~\ref{table:data_augmentation},
our system can improve accuracy by doubling the number of pairs of stored feature maps and labels. 
In the results for CIFAR-10 and STL-10, data augmentation did not always improve accuracy, but the accuracy was already high enough without data augmentation (the accuracy is higher than 94\%).

\begin{figure}[t]
   \begin{center}
   \includegraphics[width=0.8\linewidth]{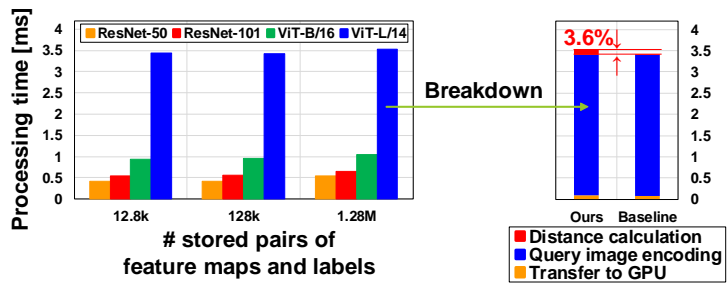}
   \end{center}
   \caption{Processing time for inference of the ImageNet-1k images as measured on an NVIDIA A100 GPU (left)
   and the breakdown comparison for ViT-L/14 with 1.28M stored pairs of feature maps and labels (right).
   The baseline is the processing time of the zero-shot CLIP classifier.}
   \label{fig:inference_time_breakdown}
\end{figure}

Fig.~\ref{fig:inference_time_breakdown} shows the processing time for inference of the ImageNet-1k images that was measured on the NVIDIA A100 GPU.
The left graph shows the processing time of our system when storing 12.8k, 128k, and 1.28M pairs of feature maps and labels with different image encoder models.
In this experiment, the maximum required memory capacity is 5.3 GB when using 1.28M pairs of feature maps and labels, and all the feature maps can be fully loaded into GPU memory.
By exploiting massive parallel computations of GPU, we execute a linear search based on cosine distance between a query and all the feature maps calculated in Eq.~\eqref{eq:cos_sim}.
As shown in the left graph, as the number of stored pairs increases, the processing time becomes slightly longer (e.g., 3\% in the case of ViT-L/14).
The right graph in Fig.~\ref{fig:inference_time_breakdown} is the breakdown of the processing time for ViT-L/14 with 1.28M stored pairs.
For comparison, the right graph also includes the breakdown of the zero-shot CLIP classifier as a baseline.
The processing time for query image encoding is dominant, but that for the distance calculation is short enough (about 0.1 ms).
The overhead from the baseline for retrieving similar feature maps from 1.28M stored ones is as small as 3.6\%.

\subsection{Continual Learning}
\label{exprmnt:continual_learn}

\noindent
\textbf{Experimental setup.} We evaluated the applicability of our system to continual learning, especially in task incremental and class incremental settings.

Task incremental learning is a setting where the number of tasks increases step by step.
We used a Split CIFAR-100 dataset~\cite{AGEM,SplitCIFAR100} in which the CIFAR-100 dataset is split into 20 disjoint subsets.
Each subset consists of five randomly sampled classes without duplication from a total of 100 classes.
Additionally, we applied ImageNet-100~\cite{ImageNet100,ImageNet100_2,PODNet} and ImageNet-1k datasets to the incremental learning setting.
ImageNet-100 consists of 100 classes randomly sampled from 1000 classes in the ImageNet-1k dataset.
We split the ImageNet-100 and ImageNet-1k datasets into 10 disjoint subsets, each containing 10 and 100 classes, respectively.
We did not balance the number of samples in each subset.
We used each subset in Split CIFAR-100, Split ImageNet-100, and Split ImageNet-1k as one task for a 5-class, 10-class, and 100-class classification, respectively.

Class incremental learning is a setting in which the number of classes is continually added.
In our experiment, we increased the number of classes by 5, 10, and 100 by sequentially adding the classes in each subset of Split CIFAR-100, Split ImageNet-100, and Split ImageNet-1k, respectively.

For each setting,
we used ResNet-50, ViT-B/32, and ViT-L/14 models for the image encoder, and employed the pretrained models by CLIP as Section~\ref{exprmnt:basic_evaluation}.
We evaluated the accuracy of our system, the zero-shot CLIP classifier,
and the conventional methods proposed for the incremental learning settings (iCaRL, ER, GEM, A-GEM, and EWC~\cite{iCaRL,ER,GEM,AGEM,EWC}).
We applied the same pretrained models to the initial values for all the methods.
In iCaRL, ER, GEM, and A-GEM, a part of training data learned in the past is stored in memory, and when new data is added, the model parameters are updated along with the stored data.
EWC does not store any training data learned in the past. 
The detailed conditions such as the hyperparameter settings are described in the supplementary materials.\\

\begin{figure}[t]
   \begin{center}
   \includegraphics[width=0.9\linewidth]{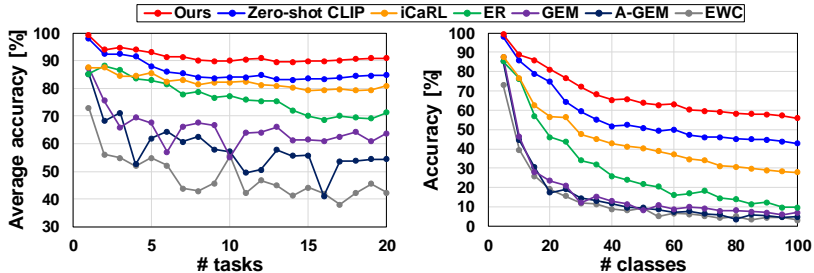}
   \end{center}
   \caption{Accuracy curves on the Split CIFAR-100 dataset in the incremental learning settings (left: task incremental, right: class incremental).}
   \label{fig:accuracy_inclearn}
\end{figure}

\noindent
\textbf{Experimental results.} The left graph in Fig.~\ref{fig:accuracy_inclearn} shows the average accuracy on the Split CIFAR-100 dataset for task incremental learning.
The average accuracy is calculated by averaging the test accuracy evaluated in each task. 
In the conventional methods (except the zero-shot CLIP classifier),
as the number of tasks increases, the average accuracy over tasks gradually drops because the models forget the knowledge for older tasks.
On the other hand, our system stores extracted image feature maps and labels, so it does not require fine-tuning to acquire new knowledge.
Therefore, our system avoids catastrophic forgetting and keeps the average accuracy over 90\%, even when the number of tasks exceeds 15.
The right graph in Fig.~\ref{fig:accuracy_inclearn} shows the accuracy for class incremental learning.
The accuracy curve of our system is higher than those of the conventional methods, even as the number of classes increases.

Table~\ref{table:accuracy_inclearn} summarizes the average accuracy on the Split CIFAR-100, Split ImageNet-100, and Split ImageNet-1k datasets.
The accuracy of task incremental learning is the average of the accuracy evaluated on each task after all tasks have been added.
The accuracy of class incremental learning is the average of the each accuracy from the first to the last class setting, which is introduced in Ref.~\cite{iCaRL}
(e.g., the average of the accuracies when the number of classes is from 5 to 100 in the right figure of Fig.~\ref{fig:accuracy_inclearn}).
Our system achieves better accuracy than the conventional methods, which indicates that our system is effective in incremental learning settings.

\begin{table}[t]
\centering
\caption{Accuracy evaluations in task and class incremental learning settings.}
\begin{minipage}{0.45\textwidth}
\centering
\scalebox{0.85}{
\begin{tabular}{ccc} 
\multicolumn{3}{c}{ResNet-50 on Split CIFAR-100} \\ \hline
\textbf{Method} & \textbf{task} & \textbf{class} \\ \hline
EWC~\cite{EWC}             & 42.2 & 13.5 \\
A-GEM~\cite{AGEM}          & 54.5 & 15.8 \\
GEM~\cite{GEM}             & 63.7 & 17.4 \\
ER~\cite{ER}               & 71.3 & 29.4 \\
iCaRL~\cite{iCaRL}         & 80.9 & 43.9 \\
Zero-shot CLIP~\cite{CLIP} & 84.8 & 56.6 \\ \hline
\textbf{Ours}              & \textbf{90.8} & \textbf{67.9} \\ \hline
\end{tabular}
}
\end{minipage}
\begin{minipage}{0.45\textwidth}
\centering
\scalebox{0.85}{
\begin{tabular}{ccc} 
\multicolumn{3}{c}{ViT-B/32 on Split ImageNet-100} \\ \hline
\textbf{Method} & \textbf{task} & \textbf{class} \\ \hline
Zero-shot CLIP~\cite{CLIP} & 92.0 & 80.2 \\
\textbf{Ours}              & \textbf{94.3} & \textbf{85.1} \\ \hline
\end{tabular}
}
\scalebox{0.85}{
\begin{tabular}{ccc} \\
\multicolumn{3}{c}{ViT-L/14 on Split ImageNet-1k} \\ \hline
\textbf{Method} & \textbf{task} & \textbf{class} \\ \hline
Zero-shot CLIP~\cite{CLIP} & 93.0 & 82.3 \\
\textbf{Ours}              & \textbf{94.2} & \textbf{85.5} \\ \hline
\end{tabular}
}
\end{minipage}
\label{table:accuracy_inclearn}
\end{table}

\subsection{Correcting false knowledge}
\noindent
\textbf{Experimental setup.} Datasets used as support sets can contain incorrectly labeled images.
In this section, we observe the impact of incorrectly labeled images on classification results and accuracy, and evaluate the effect of eliminating them.
According to Refs.~\cite{ImageNetLabelError,ImageNetReaL}, the ImageNet-1k dataset contains a small number of incorrectly labeled images (at least 6\%).
Ref.~\cite{ImageNetReaL} reassesses the label of each image in ImageNet-1k and releases an ImageNet-ReaL dataset
that eliminates the incorrectly labeled images from the original ImageNet-1k dataset.
We compared the accuracy before and after eliminating the incorrectly labeled images from the training dataset, namely, the support set.\\

\begin{figure}[t]
   \begin{center}
   \includegraphics[width=0.9\linewidth]{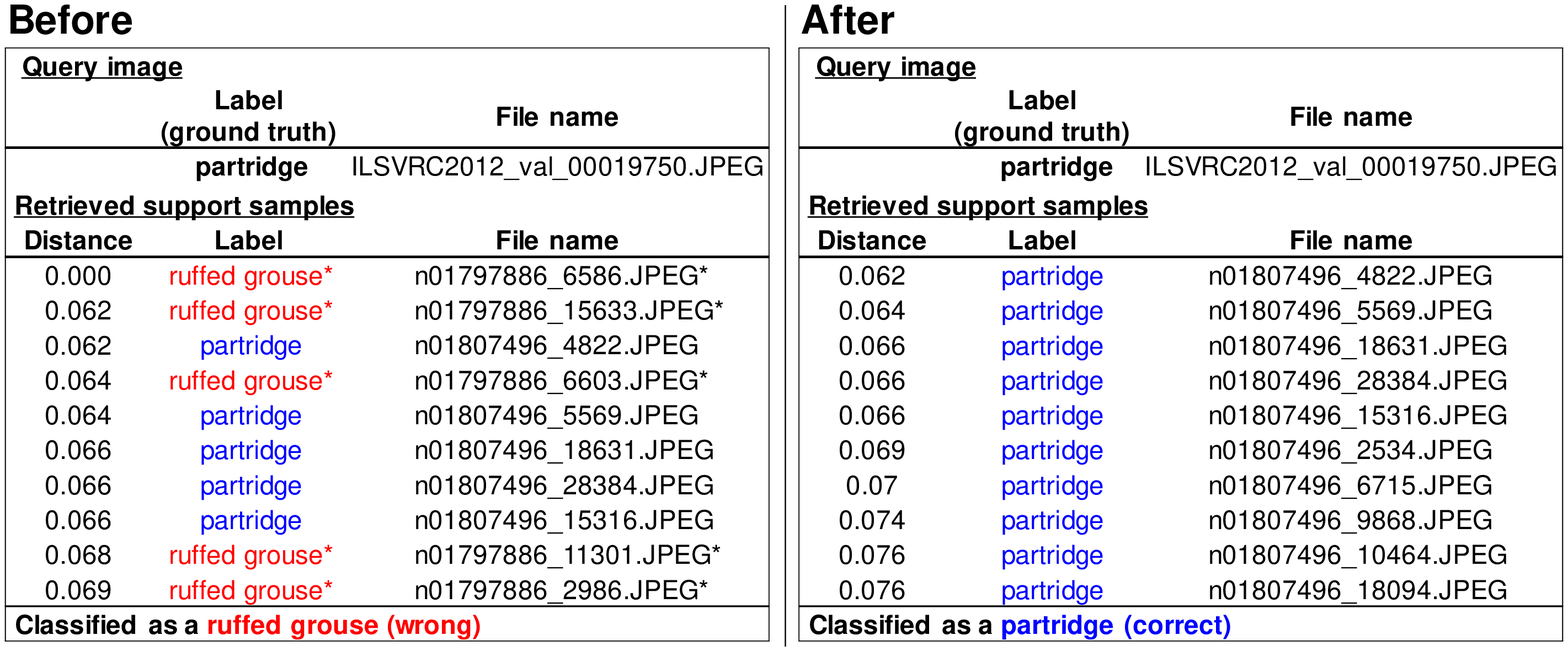}
   \end{center}
   \caption{Examples of query and top-10 similar support samples retrieved from ImageNet-1k before and after eliminating the incorrectly labeled samples from the support set.
   Before the elimination, the query image is \textcolor{red}{misclassified as a ruffed grouse} due to *incorrectly labeled samples.
   After the elimination, the query image is \textcolor{blue}{correctly classified as a partridge} in all samples.}
   \label{fig:retrieved_features_eliminate}
\end{figure}

\noindent
\textbf{Experimental results.}
Fig.~\ref{fig:retrieved_features_eliminate} shows an example of samples retrieved with the ViT-L/14 model before and after eliminating incorrectly labeled images from the support set.
Before the elimination, a query image is misclassified as a ruffed grouse (correctly as a partridge) due to the retrieval of feature maps of incorrectly labeled images.
After the elimination, the incorrectly labeled samples are no longer included in the top-10 retrieved samples, and the query image is correctly classified as a partridge.
These verification results of Fig.~\ref{fig:retrieved_features_eliminate} can be visualized with the support images (like Fig.~\ref{fig:retrieved_features} in Section~\ref{sec:pm_storing}).
In this experiment, when all the support images in the ImageNet-1k dataset are stored for the verification, the required storage capacity is 152GB.

Table~\ref{table:imagenet_real} shows the accuracy on the ImageNet-ReaL dataset before and after the elimination.
Our system can correct false knowledge and improve accuracy by simply eliminating feature maps of incorrectly labeled images.
Unlike existing image classification systems, our system does not need to fine-tune model parameters again on the modified dataset.

\begin{table}[t]
\caption{\textbf{Left}: validation accuracy on the ImageNet-ReaL dataset before and after eliminating incorrectly labeled images from the support sets.
         \textbf{Right}: the details for the support and query sets.}
\begin{minipage}{0.45\textwidth}
\centering
\scalebox{0.8}{
\begin{tabular}{ccc|ccc|}
\hline 
\multicolumn{1}{|c}{}&\multicolumn{2}{|c}{ResNet}&\multicolumn{3}{|c|}{ViT} \\ 
\multicolumn{1}{|c}{}&\multicolumn{1}{|c}{50}    &  101  & B/32 & B/16 & L/14 \\ \hline 
\multicolumn{1}{|c}{\textbf{Before}}     &\multicolumn{1}{|c}{71.2} & 75.7  & 75.0 & 79.3 & 83.9 \\
\multicolumn{1}{|c}{\textbf{After}}     &\multicolumn{1}{|c}{71.4} & 76.1  & 75.4 & 79.6 & 84.0 \\
\multicolumn{1}{|c}{$\Delta$} &\multicolumn{1}{|c}{\textbf{+0.2}} & \textbf{+0.4} & \textbf{+0.4} & \textbf{+0.3} & \textbf{+0.1} \\ \hline
\end{tabular}
}
\end{minipage}
\label{table:imagenet_real}
\begin{minipage}{0.45\textwidth}
\centering
\scalebox{0.8}{
\begin{tabular}{|c|c|c|}
\hline 
& \textbf{Support set} & \textbf{Query set} \\ \hline
\multirow{2}{*}{\textbf{Before}} & 1,281,167 images & 46,837 images \\
& in ImageNet-1k & in ImageNet-ReaL \\ \hline
\multirow{2}{*}{\textbf{After}}  & 1,148,659 images & 46,837 images \\
& in ImageNet-ReaL & in ImageNet-ReaL \\ \hline
\end{tabular}
}
\end{minipage}
\end{table}

\section{Conclusions}

In this paper, we investigated a system that stores knowledge for image classification not in model parameters but in external high-capacity storage
and refers to this storage like a database when classifying input images.
Our system can increase knowledge by adding image feature maps and their corresponding labels to the database.
Without fine-tuning the model parameters, our system avoids catastrophic forgetting in continual learning scenarios
and achieves better accuracy than the conventional methods.
By reviewing the neighborhood samples retrieved by the kNN classifier, we can verify the classification results.
If incorrectly labeled samples are included in the classification results,
then our system can correct the false knowledge and improve accuracy without fine-tuning by simply eliminating the incorrectly labeled samples from the database.
In future work, we will work on scaling up our system with high-capacity storage
by utilizing a large-scale fast ANN (approximate nearest neighbor) search.

\clearpage
\bibliographystyle{splncs04}
\bibliography{egbib}

\begin{thebibliography}{10}
\providecommand{\url}[1]{\texttt{#1}}
\providecommand{\urlprefix}{URL }
\providecommand{\doi}[1]{https://doi.org/#1}

\bibitem{ImageNetReaL}
Beyer, L., H{\'{e}}naff, O.J., Kolesnikov, A., Zhai, X., van~den Oord, A.: {Are
  we done with ImageNet?} arXiv  \textbf{abs/2006.07159} (2020)

\bibitem{foundation_model}
Bommasani, R., Hudson, D.A., Adeli, E., Altman, R., Arora, S., von Arx, S.,
  Bernstein, M.S., Bohg, J., Bosselut, A., Brunskill, E., Brynjolfsson, E.,
  Buch, S., Card, D., Castellon, R., Chatterji, N.S., Chen, A.S., Creel, K.,
  Davis, J.Q., Demszky, D., Donahue, C., Doumbouya, M., Durmus, E., Ermon, S.,
  Etchemendy, J., Ethayarajh, K., Fei{-}Fei, L., Finn, C., Gale, T., Gillespie,
  L., Goel, K., Goodman, N.D., Grossman, S., Guha, N., Hashimoto, T.,
  Henderson, P., Hewitt, J., Ho, D.E., Hong, J., Hsu, K., Huang, J., Icard, T.,
  Jain, S., Jurafsky, D., Kalluri, P., Karamcheti, S., Keeling, G., Khani, F.,
  Khattab, O., Koh, P.W., Krass, M.S., Krishna, R., Kuditipudi, R., et~al.: {On
  the Opportunities and Risks of Foundation Models}. arXiv
  \textbf{abs/2108.07258} (2021)

\bibitem{RETRO}
Borgeaud, S., Mensch, A., Hoffmann, J., Cai, T., Rutherford, E., Millican, K.,
  van~den Driessche, G., Lespiau, J., Damoc, B., Clark, A., de~Las~Casas, D.,
  Guy, A., Menick, J., Ring, R., Hennigan, T., Huang, S., Maggiore, L., Jones,
  C., Cassirer, A., Brock, A., Paganini, M., Irving, G., Vinyals, O., Osindero,
  S., Simonyan, K., Rae, J.W., Elsen, E., Sifre, L.: {Improving language models
  by retrieving from trillions of tokens}. arXiv  \textbf{abs/2112.04426}
  (2021)

\bibitem{DARK_ER}
Buzzega, P., Boschini, M., Porrello, A., Abati, D., CALDERARA, S.: {Dark
  Experience for General Continual Learning: a Strong, Simple Baseline}. In:
  NeurIPS (2020)

\bibitem{ERTrick}
Buzzega, P., Boschini, M., Porrello, A., Calderara, S.: {Rethinking Experience
  Replay: a Bag of Tricks for Continual Learning}. In: ICPR (2021)

\bibitem{DINO}
Caron, M., Touvron, H., Misra, I., J{\'{e}}gou, H., Mairal, J., Bojanowski, P.,
  Joulin, A.: {Emerging Properties in Self-Supervised Vision Transformers}.
  arXiv  \textbf{abs/2104.14294} (2021)

\bibitem{HAL}
Chaudhry, A., Gordo, A., Dokania, P.K., Torr, P.H.S., Lopez-Paz, D.: {Using
  Hindsight to Anchor Past Knowledge in Continual Learning}. In: AAAI (2021)

\bibitem{AGEM}
Chaudhry, A., Ranzato, M., Rohrbach, M., Elhoseiny, M.: {Efficient Lifelong
  Learning with A-GEM}. In: ICLR (2019)

\bibitem{TIBAV}
Chefer, H., Gur, S., Wolf, L.: {Transformer Interpretability Beyond Attention
  Visualization}. In: CVPR (2021)

\bibitem{SPANN}
Chen, Q., Zhao, B., Wang, H., Li, M., Liu, C., Li, Z., Yang, M., Wang, J.:
  {SPANN: Highly-efficient Billion-scale Approximate Nearest Neighborhood
  Search}. In: NeurIPS (2021)

\bibitem{SimCLR}
Chen, T., Kornblith, S., Norouzi, M., Hinton, G.E.: {A Simple Framework for
  Contrastive Learning of Visual Representations}. arXiv
  \textbf{abs/2002.05709} (2020)

\bibitem{MOCOV2}
Chen, X., Fan, H., Girshick, R.B., He, K.: {Improved Baselines with Momentum
  Contrastive Learning}. arXiv  \textbf{abs/2003.04297} (2020)

\bibitem{STL10}
Coates, A., Ng, A., Lee, H.: {An Analysis of Single-Layer Networks in
  Unsupervised Feature Learning}. In: AISTATS (2011)

\bibitem{kNNreview}
Cunningham, P., Delany, S.J.: {k-Nearest Neighbour Classifiers: 2nd Edition
  (with Python examples)}. arXiv  \textbf{abs/2004.04523} (2020)

\bibitem{CoAtNet}
Dai, Z., Liu, H., Le, Q.V., Tan, M.: {CoAtNet: Marrying Convolution and
  Attention for All Data Sizes}. arXiv  \textbf{abs/2106.04803} (2021)

\bibitem{ImageNet}
Deng, J., Dong, W., Socher, R., Li, L.J., Li, K., Fei-Fei, L.: {ImageNet: A
  large-scale hierarchical image database}. In: CVPR (2009)

\bibitem{ViT}
Dosovitskiy, A., Beyer, L., Kolesnikov, A., Weissenborn, D., Zhai, X.,
  Unterthiner, T., Dehghani, M., Minderer, M., Heigold, G., Gelly, S.,
  Uszkoreit, J., Houlsby, N.: {An Image is Worth 16x16 Words: Transformers for
  Image Recognition at Scale}. In: ICLR (2021)

\bibitem{PODNet}
Douillard, A., Cord, M., Ollion, C., Robert, T., Valle, E.: {PODNet: Pooled
  Outputs Distillation for Small-Tasks Incremental Learning}. In: ECCV (2020)

\bibitem{NNCLR}
Dwibedi, D., Aytar, Y., Tompson, J., Sermanet, P., Zisserman, A.: {With a
  Little Help From My Friends: Nearest-Neighbor Contrastive Learning of Visual
  Representations}. In: ICCV (2021)

\bibitem{cataforget}
French, R.M.: {Catastrophic forgetting in connectionist networks}. Trends in
  Cognitive Sciences  \textbf{3} (1999)

\bibitem{CCT}
Hassani, A., Walton, S., Shah, N., Abuduweili, A., Li, J., Shi, H.: {Escaping
  the Big Data Paradigm with Compact Transformers}. arXiv
  \textbf{abs/2104.05704} (2021)

\bibitem{MAE}
He, K., Chen, X., Xie, S., Li, Y., Doll{\'{a}}r, P., Girshick, R.B.: {Masked
  Autoencoders Are Scalable Vision Learners}. arXiv  \textbf{abs/2111.06377}
  (2021)

\bibitem{MOCO}
He, K., Fan, H., Wu, Y., Xie, S., Girshick, R.: {Momentum Contrast for
  Unsupervised Visual Representation Learning}. In: CVPR (2020)

\bibitem{ResNet}
He, K., Zhang, X., Ren, S., Sun, J.: {Deep Residual Learning for Image
  Recognition}. In: CVPR (2016)

\bibitem{SurveyOnlineLearn}
Hoi, S.C., Sahoo, D., Lu, J., Zhao, P.: {Online learning: A comprehensive
  survey}. Neurocomputing  \textbf{459} (2021)

\bibitem{ImageNet100}
Hou, S., Pan, X., Loy, C.C., Wang, Z., Lin, D.: {Learning a Unified Classifier
  Incrementally via Rebalancing}. In: CVPR (2019)

\bibitem{ALIGN}
Jia, C., Yang, Y., Xia, Y., Chen, Y.T., Parekh, Z., Pham, H., Le, Q., Sung,
  Y.H., Li, Z., Duerig, T.: {Scaling Up Visual and Vision-Language
  Representation Learning With Noisy Text Supervision}. In: ICML (2021)

\bibitem{FAISS}
Johnson, J., Douze, M., J{\'e}gou, H.: {Billion-scale similarity search with
  GPUs}. IEEE Transactions on Big Data  \textbf{7}(3),  535--547 (2019)

\bibitem{DPR}
Karpukhin, V., Oguz, B., Min, S., Lewis, P., Wu, L., Edunov, S., Chen, D., Yih,
  W.t.: {Dense Passage Retrieval for Open-Domain Question Answering}. In: EMNLP
  (2020)

\bibitem{EWC}
Kirkpatrick, J., Pascanu, R., Rabinowitz, N., Veness, J., Desjardins, G., Rusu,
  A.A., Milan, K., Quan, J., Ramalho, T., Grabska-Barwinska, A., Hassabis, D.,
  Clopath, C., Kumaran, D., Hadsell, R.: {Overcoming catastrophic forgetting in
  neural networks}. Proceedings of the National Academy of Sciences
  \textbf{114} (2017)

\bibitem{datapoints_attention}
Kossen, J., Band, N., Lyle, C., Gomez, A., Rainforth, T., Gal, Y.:
  {Self-Attention Between Datapoints: Going Beyond Individual Input-Output
  Pairs in Deep Learning}. In: NeurIPS (2021)

\bibitem{CIFAR10}
Krizhevsky, A.: {Learning multiple layers of features from tiny images}. Tech.
  rep. (2009)

\bibitem{AlexNet}
Krizhevsky, A., Sutskever, I., Hinton, G.E.: {ImageNet Classification with Deep
  Convolutional Neural Networks}. In: NIPS (2012)

\bibitem{SurveyIncLearn}
Lange, M.D., Aljundi, R., Masana, M., Parisot, S., Jia, X., Leonardis, A.,
  Slabaugh, G.G., Tuytelaars, T.: {Continual learning: A comparative study on
  how to defy forgetting in classification tasks}. arXiv
  \textbf{abs/1909.08383} (2019)

\bibitem{LeNet}
Lecun, Y., Bottou, L., Bengio, Y., Haffner, P.: {Gradient-based learning
  applied to document recognition}. Proceedings of the IEEE  \textbf{86} (1998)

\bibitem{EsViT}
Li, C., Yang, J., Zhang, P., Gao, M., Xiao, B., Dai, X., Yuan, L., Gao, J.:
  {Efficient Self-supervised Vision Transformers for Representation Learning}.
  In: ICLR (2022)

\bibitem{longtail_retrieval}
Long, A., Yin, W., Ajanthan, T., Nguyen, V., Purkait, P., Garg, R., Blair, A.,
  Shen, C., van~den Hengel, A.: {Retrieval Augmented Classification for
  Long-Tail Visual Recognition}. arXiv  \textbf{abs/2202.11233} (2022)

\bibitem{GEM}
Lopez-Paz, D., Ranzato, M.: {Gradient Episodic Memory for Continual Learning}.
  In: NIPS (2017)

\bibitem{ImageNetLabelError}
Northcutt, C.G., Athalye, A., Mueller, J.: {Pervasive Label Errors in Test Sets
  Destabilize Machine Learning Benchmarks}. In: NeurIPS Datasets and Benchmarks
  Track (2021)

\bibitem{pytorch}
Paszke, A., Gross, S., Chintala, S., Chanan, G., Yang, E., DeVito, Z., Lin, Z.,
  Desmaison, A., Antiga, L., Lerer, A.: {Automatic Differentiation in PyTorch}.
  In: NIPS 2017 Workshop on Autodiff (2017)

\bibitem{CLIP}
Radford, A., Kim, J.W., Hallacy, C., Ramesh, A., Goh, G., Agarwal, S., Sastry,
  G., Askell, A., Mishkin, P., Clark, J., Krueger, G., Sutskever, I.: {Learning
  Transferable Visual Models From Natural Language Supervision}. In: ICML
  (2021)

\bibitem{iCaRL}
Rebuffi, S.A., Kolesnikov, A., Sperl, G., Lampert, C.H.: {iCaRL: Incremental
  Classifier and Representation Learning}. In: CVPR (2017)

\bibitem{ER}
Rolnick, D., Ahuja, A., Schwarz, J., Lillicrap, T., Wayne, G.: {Experience
  Replay for Continual Learning}. In: NeurIPS (2019)

\bibitem{GradCAM}
Selvaraju, R.R., Cogswell, M., Das, A., Vedantam, R., Parikh, D., Batra, D.:
  {Grad-CAM: Visual Explanations From Deep Networks via Gradient-Based
  Localization}. In: ICCV (2017)

\bibitem{VGG}
Simonyan, K., Zisserman, A.: {Very Deep Convolutional Networks for Large-Scale
  Image Recognition}. In: ICLR (2015)

\bibitem{UncoveringBiases}
Stock, P., Cisse, M.: {ConvNets and ImageNet Beyond Accuracy: Understanding
  Mistakes and Uncovering Biases}. In: ECCV (2018)

\bibitem{CommonsenseQA}
Talmor, A., Herzig, J., Lourie, N., Berant, J.: {CommonsenseQA: A Question
  Answering Challenge Targeting Commonsense Knowledge}. In: NAACL-HLT (2019)

\bibitem{CompStorage}
Torabzadehkashi, M., Rezaei, S., HeydariGorji, A., Bobarshad, H., Alves, V.,
  Bagherzadeh, N.: {Computational storage: an efficient and scalable platform
  for big data and hpc applications}. Journal of Big Data  \textbf{6}(1),
  1--29 (2019)

\bibitem{Transformer}
Vaswani, A., Shazeer, N., Parmar, N., Uszkoreit, J., Jones, L., Gomez, A.N.,
  Kaiser, L.u., Polosukhin, I.: {Attention is All you Need}. In: NIPS (2017)

\bibitem{PyramidViT}
Wang, W., Xie, E., Li, X., Fan, D., Song, K., Liang, D., Lu, T., Luo, P., Shao,
  L.: {Pyramid Vision Transformer: A Versatile Backbone for Dense Prediction
  without Convolutions}. arXiv  \textbf{abs/2102.12122} (2021)

\bibitem{CLD}
Wang, X., Liu, Z., Yu, S.X.: {Unsupervised Feature Learning by Cross-Level
  Instance-Group Discrimination}. In: CVPR (2021)

\bibitem{AdvPrune}
Wu, D., Wang, Y.: {Adversarial Neuron Pruning Purifies Backdoored Deep Models}.
  In: NeurIPS (2021)

\bibitem{ImageNet100_2}
Wu, Y., Chen, Y., Wang, L., Ye, Y., Liu, Z., Guo, Y., Fu, Y.: {Large Scale
  Incremental Learning}. In: CVPR (2019)

\bibitem{ID}
Wu, Z., Xiong, Y., Yu, S.X., Lin, D.: {Unsupervised Feature Learning via
  Non-parametric Instance Discrimination}. In: CVPR (2018)

\bibitem{hummancommonsense}
Xu, Y., Zhu, C., Wang, S., Sun, S., Cheng, H., Liu, X., Gao, J., He, P., Zeng,
  M., Huang, X.: {Human Parity on CommonsenseQA: Augmenting Self-Attention with
  External Attention}. arXiv  \textbf{abs/2112.03254} (2021)

\bibitem{SplitCIFAR100}
Zenke, F., Poole, B., Ganguli, S.: {Continual Learning Through Synaptic
  Intelligence}. In: ICML (2017)

\bibitem{reinforcement_pruning}
Zhong, J., Ding, G., Guo, Y., Han, J., Wang, B.: {Where to Prune: Using LSTM to
  Guide End-to-end Pruning}. In: IJCAI (2018)

\bibitem{CAM}
Zhou, B., Khosla, A., Lapedriza, {\`A}., Oliva, A., Torralba, A.: {Learning
  Deep Features for Discriminative Localization}. In: CVPR (2016)

\bibitem{iBOT}
Zhou, J., Wei, C., Wang, H., Shen, W., Xie, C., Yuille, A., Kong, T.: {Image
  BERT Pre-training with Online Tokenizer}. In: ICLR (2022)

\bibitem{bayesian_pruning}
Zhou, Y., Zhang, Y., Wang, Y., Tian, Q.: {Accelerate CNN via Recursive Bayesian
  Pruning}. In: ICCV (2019)

\end{thebibliography}

\newpage

\appendix

\section*{Appendix}

\section{Applications of our system}
Our system registers feature maps extracted from images and their corresponding labels to the database, as described in Section~\ref{sec:pm_storing}.
A feature map can be linked with not only a single label but multiple labels as well.
For example, when extracting a feature map from a \textit{cat} image, our system can register
not only the \textit{cat} label but also the labels \textit{tabby} (the subcategory) and \textit{mammal} or \textit{animal}
(the broad category or superclass) with the feature map.
In this work, we introduced the procedure of linear search for all feature maps in the storage
when calculating the distance with the feature maps of the query image.
Our system can use multiple registered labels to refine the search according to the classification setting.
For instance, in the case of an animal classification problem, our system can selectively list
the files of image feature maps linked to the \textit{animal} label before the distance calculation.
Then, by loading only the listed image feature maps to the computing device,
the distance calculation can be processed even with limited memory resources.

By analyzing the number of references for each sample in the database, we can identify frequently or rarely referred samples, and interpret what knowledge is currently required or unnecessary.
The information on the number of references can be used for prioritization when verifying the validity of stored knowledge.
For example, when checking labels of samples within a limited amount of time, users can review the labels of frequently referred samples first, instead of checking all the samples in the storage.
% by sorting the number of references in descending order,
Moreover, if users need to reduce the storage size, rarely referred samples can be deleted from the storage like pruning model parameters.

In addition, the inference time of our system is as short as a few milliseconds, as shown in Fig.~\ref{fig:inference_time_breakdown}.
Therefore, we believe that our system can be implemented in real-time applications,
e.g., infrastructure monitoring systems,
where objects to be detected change temporally and need to be identified in real time without training.
Our system can log the samples referred for inference, along with the original images and file names.
When some anomaly or error occurs, operators can effectively investigate the cause later by checking the logs.

\section{Pretrained image encoder models}
\label{sec:details_image_encoders}

In this work, we primarily employed image encoder models pretrained by CLIP~\cite{CLIP}. 
We summarize the details of the image encoders in Table~\ref{table:detail_image_encoder}.

\begin{table}[t]
\begin{center}
\caption{Details of image encoder models.}
\begin{tabular}{cccccc}
\hline 
& \textbf{Output feature} & \textbf{Input} & \multicolumn{2}{r}{\textbf{ResNet}} &\\
\hspace{1.5pt} \textbf{Model} \hspace{1.5pt}& \hspace{1.5pt} \textbf{dimension} \hspace{1.5pt} & \hspace{1.5pt} \textbf{resolution} \hspace{0.5pt} & \multicolumn{2}{c}{\hspace{0.5pt}\textbf{blocks}\hspace{1.5pt}} & \hspace{1.5pt} \textbf{width} \hspace{1.5pt} \\ \hline 
RN50 & 1024 & 224 & \multicolumn{2}{c}{\hspace{2pt}(3,4,6,3)\hspace{2pt}} & \hspace{1.5pt}2048\hspace{1.5pt} \\
RN101 & 512 & 224 & \multicolumn{2}{c}{\hspace{2pt}(3,4,23,3)\hspace{2pt}} & \hspace{1.5pt}2048\hspace{1.5pt} \\ \hline
\\ \hline
& \textbf{Output feature} & \textbf{Input} & \multicolumn{3}{c}{\textbf{Vision Transformer}} \\
\hspace{1.5pt} \textbf{Model} \hspace{1.5pt}& \hspace{1.5pt} \textbf{dimension} \hspace{1.5pt}& \hspace{1.5pt} \textbf{resolution} \hspace{0.5pt} & \hspace{0.5pt} \textbf{layers} \hspace{1.5pt} & \hspace{1.5pt} \textbf{width} \hspace{1.5pt} & \hspace{1.5pt} \textbf{head} \hspace{1.5pt}\\ \hline 
ViT-B/32 & 512 & 224 & \hspace{1pt}12\hspace{1pt} & \hspace{1pt}768\hspace{1pt} & \hspace{1pt}12\hspace{1pt} \\
ViT-B/16 & 512 & 224 & \hspace{1pt}12\hspace{1pt} & \hspace{1pt}768\hspace{1pt} & \hspace{1pt}12\hspace{1pt} \\
ViT-L/14 & 768 & 224 & \hspace{1pt}24\hspace{1pt} & \hspace{1pt}1024\hspace{1pt} & \hspace{1pt}16\hspace{1pt} \\ \hline

\end{tabular}
\label{table:detail_image_encoder}
\end{center}
\end{table}

\section{Processing time for distance calculation}

\begin{figure}[h]
   \begin{center}
   \includegraphics[width=0.65\linewidth]{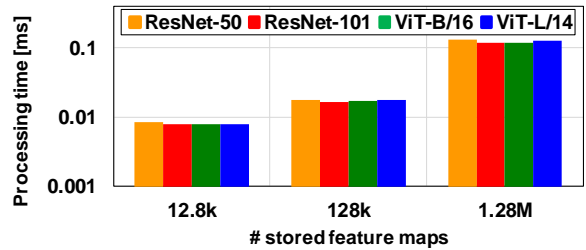}
   \end{center}
   \caption{Processing time for distance calculation measured on the NVIDIA A100 GPU.}
   \label{fig_process_time}
\end{figure}

As described in Section~\ref{sec:pm_inference}, our system retrieves nearest neighbor samples based on the cosine distance between a query and the stored feature maps.
In the experiments in this work, the memory capacity required for using the stored feature maps is 5.3 GB at most.
Therefore, all of them can be fully loaded into GPU memory, and a linear search based on the distance calculation is executed on GPU.
Fig.~\ref{fig_process_time} shows the processing time for distance calculation between a query and the stored feature maps measured on the NVIDIA A100 GPU.
As shown in Fig.~\ref{fig_process_time}, the processing time increases with the number of stored feature maps.
However, in terms of the overall inference procedure, the processing time for query encoding is more dominant than that for distance calculation, 
so the total processing time does not increase much as the number of stored feature maps increases, as shown in Fig.~\ref{fig:inference_time_breakdown}.

If the number of stored feature maps is scaled up (e.g., 10$\times$), all the feature maps may not fit completely into the memory.
Then, the time required for memory access will be non-negligible.
As methods for scaling up our system, we can utilize
(i)~large-scale fast ANN (approximate nearest neighbor) search with storage~\cite{SPANN,FAISS},
(ii)~larger memory with the latest GPUs, and/or (iii)~computational storage~\cite{CompStorage},
and we are currently investigating them as future work.

\section{Hyperparameter settings in continual learning scenarios}
\label{sec:hyperparam}

In Section~\ref{exprmnt:continual_learn}, we compared the performance of our system with those of conventional methods proposed for continual learning scenarios (iCaRL, ER, GEM, A-GEM, and EWC~\cite{iCaRL,ER,GEM,AGEM,EWC}).
We employed a fully connected layer as a classifier head for these methods, and applied pretrained models by CLIP to the initial values.
We employed a stochastic gradient descent algorithm as an optimizer and trained the image encoder and classifier models for 50 epochs on the Split CIFAR-100 dataset~\cite{AGEM,SplitCIFAR100}.
We set the learning rate to 0.01 and the mini-batch size to 32.
In iCaRL, ER, GEM, and A-GEM, we set memory buffer size to 500 for storing a part of training data learned in the past in the memory.
In EWC, we set $\lambda$ for the penalty term to 0.5 without storing any training data learned in the past.
For these hyperparameter settings, we referred to the values in Refs.~\cite{DARK_ER,ERTrick}.

\section{Final accuracy evaluation in class incremental setting} 

In Table~\ref{table:accuracy_inclearn} in Section~\ref{exprmnt:continual_learn}, we evaluated the performance in class-incremental learning with the averaged accuracy values from the first to the last set of classes.
We can also evaluate the performance with the final accuracy values after the last set of classes has been learned.
Table~\ref{table:accuracy_inclearn} summarizes the final accuracy in class-incremental learning with ResNet-50 on Split CIFAR-100.
As shown in Table~\ref{table:accuracy_inclearn}, our approach is also effective when evaluating the final accuracy.

\begin{table}[b]
\caption{Final accuracy evaluation in class incremental setting with ResNet-50 on Split CIFAR-100.}
\centering
\scalebox{1.0}{%0.85
\begin{tabular}{cccccccccc} 
\multicolumn{8}{c}{ResNet-50 on Split CIFAR-100} \\ \hline

\textbf{Method} & \textbf{Ours} & CLIP & iCaRL & ER & GEM & A-GEM & EWC       \\ \hline
\textbf{Accuracy [\%]} & \textbf{55.9} & 42.8 & 27.8 & 9.8 & 7.0 & 4.8 & 3.0      \\ \hline
\end{tabular}
}
\label{table:accuracy_inclearn}
\end{table}

\section{Future works for better pretraining}

In Section~\ref{sec:pm_pretraining}, we described that the performance of our system depends on the pretraining methods of image encoder models.
In this work, we directly used the image encoder models pretrained by CLIP, and we did not optimize the pretraining condition or the architecture of the pretrained models (e.g., the number of output feature dimensions) for our system.

CLIP contrastively learns image and text encoders such that relevant image-text pairs are mapped to the neighborhood in the latent space.
In this pretraining, a kNN classifier is not employed as a head model to calculate the contrastive loss, and the pretrained models are not optimized for the kNN classifier.
Recently, Ref.~\cite{NNCLR} has proposed a contrastive learning method that explicitly incorporates a nearest neighbor algorithm into the pretraining for image encoders.
This method contrastively learns image encoders so that input images and retrieved ones from a support set by a nearest neighbor search are mapped to the neighborhood in the latent space.
To further improve the accuracy of our system, we need to devise better pretraining method and condition suitable for similarity-based retrieval using kNN, which is a future work.

\end{document}